\documentclass{article}

\PassOptionsToPackage{numbers, sort&compress}{natbib}

\usepackage[preprint]{neurips_2020}

\usepackage[utf8]{inputenc} 
\usepackage[T1]{fontenc}    
\usepackage{hyperref}       
\usepackage{url}            
\usepackage{booktabs}       
\usepackage{amsfonts}       
\usepackage{nicefrac}       
\usepackage{microtype}      

\usepackage{wrapfig}
\usepackage{todonotes}
\usepackage{tabulary}
\usepackage{tabularx}
\usepackage{makecell}
\usepackage{subcaption}
\usepackage{gensymb}
\usepackage{mwe}

\usepackage{multirow} 
\usepackage{xspace}
\usepackage[capitalise]{cleveref}
\usepackage{enumitem}
\setlist[itemize]{align=parleft}

\usepackage{dsfont}
\newcommand{\kw}[1]{{\small\textsc{\MakeLowercase{#1}}}}

\newcommand{\PM}{\textsc{$PM_{2.5}$}\xspace}

\title{Dense Forecasting of Wildfire Smoke Particulate Matter Using Sparsity Invariant Convolutional Neural Networks}

\author{
  Renhao Wang \\
  Department of Computer Science\\
  University of British Columbia\\
  \texttt{renhaow@cs.ubc.ca} \\  
  \AND
  Ashutosh Bhudia \\
  Department of Physics\\
  University of British Columbia\\
  \texttt{ashu.bhudia@gmail.com} \\
  \And
  Brandon Dos Remedios \\
  Department of Physics\\
  University of British Columbia\\
  \texttt{dosremedios.brandon@gmail.com} \\
  \And
  Minnie Teng \\
  Faculty of Medicine\\
  University of British Columbia\\
  \texttt{minnie.teng@alumni.ubc.ca} \\
  \And
  Raymond Ng\\
  Department of Computer Science\\
  University of British Columbia\\
  \texttt{rng@cs.ubc.ca} \\  
}

\begin{document}

\maketitle

\begin{abstract}

Accurate forecasts of fine particulate matter (\PM) from wildfire smoke are crucial to safeguarding cardiopulmonary public health. Existing forecasting systems are trained on sparse and inaccurate ground truths, and do not take sufficient advantage of important spatial inductive biases. In this work, we present a convolutional neural network which preserves sparsity invariance throughout, and leverages multitask learning to perform dense forecasts of \PM values. We demonstrate that our model outperforms two existing smoke forecasting systems during the 2018 and 2019 wildfire season in British Columbia, Canada, predicting \PM at a grid resolution of 10 km, 24 hours in advance with high fidelity. Most interestingly, our model also generalizes to meaningful smoke dispersion patterns despite training with irregularly distributed ground truth \PM values available in only $<0.5\%$ of grid cells.

\end{abstract}

\section{Introduction}

The relentless advance of climate change has precipitated a massive increase in wildfires all over the world. The 2018 wildfire season of British Columbia, Canada, the 2019-2020 Australia bushfire season and the ongoing California wildfire season have all seen unprecedented levels of destruction. While attention has primarily been drawn to the massive loss of life and property associated with these natural disasters, smoke from wildfires can have a more insidious and longitudinal effect. The aerosols within smoke pose serious risks to cardiopulmonary health, particularly for seniors, children and those with existing health conditions. Moreover, smoke from wildfires can be carried by winds to regions thousands of kilometers away from active fires, affecting those distant regions just as severely as nearby ones. Predicting the dispersion and impact of smoke from wildfires is therefore a task of eminent importance in public health.

Smoke forecasting is often centered on predicting fine particular matter (less than 2.5 microns in diameter, or \PM) over spatial grids, up to $60$ hours in advance \citep{pavlovic2016firework, larkin2010bluesky}. Current forecasting systems are a blend of varied physical and chemical transport models, as well as statistical models \citep{larkin2010bluesky, pavlovic2016firework}. However, a number of problems plague these statistical components. Firstly, due to sparsely located air monitoring stations, ground truth \PM values used in training these components are often lacking, or filled in with mean \PM values from the closest available measurements \citep{pavlovic2016firework}. Secondly, such statistical models do not preserve spatial inductive biases, often being based on general additive models or other methods which flatten and thereby destroy the inherent spatial structure of the input data \citep{reid2015spatiotemporal, yao2018machine, zou2019machine}.

Convolutional neural networks (CNNs) can clearly be employed to address the second issue, but still suffer significantly from sparsity issues. This is the likely reason behind their general lack of use in the wildfire smoke prediction domain. \citep{larsen2020deep} circumvent this only by tackling a simpler pixel-level binary classification on the presence of smoke or not.

Our contributions are 2-fold: 

\begin{enumerate}

\item
	We present the most comprehensive wildfire smoke data ingestion framework to date, which blends existing smoke forecasting system knowledge with predictive meteorological and wildfire variables, amalgamating varied spatiotemporal data into a format amenable to convolution-based learning.
	
\item
	We propose a multiheaded convolutional neural network architecture which uses sparsity invariant layers and auxiliary autoencoding targets to overcome extreme sparsity of ground truth \PM labels. We demonstrate the viability of this network by accurately and densely forecasting \PM values for the 2018 and 2019 wildfire seasons in British Columbia, Canada, at a resolution of 10 km, and using input data available 24 hours in advance.

\end{enumerate}

\section{Method}

\subsection{Datasets}

\subsubsection{Forecasting Models}

Various deterministic and statistical models for predicting \PM exist \citep{lassman2016blending, reid2015spatiotemporal, xiao2018ensemble, zou2019machine}. Inspired by paradigms in residual learning \cite{he2016deep}, we incorporate these baseline forecasting models as inputs to our convolutional model. The idea is to allow our network to leverage prior knowledge contained within other models, such as influences of meteorological conditions, fire behavior evolution and smoke dispersion mechanics. We then simply learn a function which models potential improvements to these baselines in order to better attain the ground truth \PM values. In particular, we incorporate two prominent smoke particulate matter forecasting systems, FireWork and BlueSky Canada.

FireWork is an air quality prediction system that combines estimates from the Regional Air Quality Deterministic Prediction System (RAQDPS) and wildfire emissions \citep{pavlovic2016firework}. The latter are also estimated from hotspot and fuel consumption data obtained by the Canadian Wildfire Information System \citep{pavlovic2016firework}. The system is run twice daily, at $00$ and $12$ UTC, providing estimates of \PM at a $10$ km resolution for every hour over the next $48$ hours \citep{pavlovic2016firework}.

\begin{figure}[ht]
        \centering
        \begin{subfigure}[t]{0.475\textwidth}
            \centering
            \includegraphics[width=\textwidth]{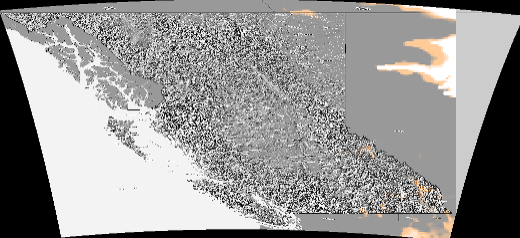}
            \caption{\footnotesize{FireWork Forecasting}}
            \label{fig:fw}
        \end{subfigure}
        \hfill
        \begin{subfigure}[t]{0.475\textwidth}
            \centering 
            \includegraphics[width=\textwidth]{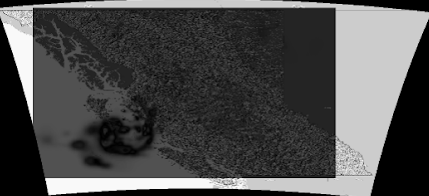}
            \caption{\footnotesize{BlueSky Canada Forecasting}}
            \label{fig:bs}
        \end{subfigure}
        \vskip \baselineskip
        \begin{subfigure}[t]{0.475\textwidth}
            \centering 
            \includegraphics[width=\textwidth]{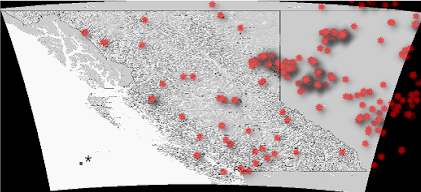}
            \caption{\footnotesize{MODIS FRP Active Fire Sites}}
            \label{fig:frp}
        \end{subfigure}
        \hfill
        \begin{subfigure}[t]{0.475\textwidth}
            \centering 
            \includegraphics[width=\textwidth]{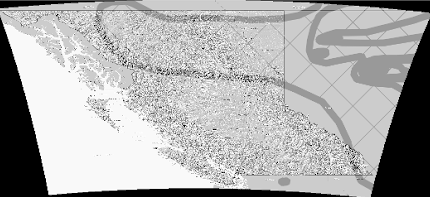}
            \caption{\footnotesize{NOAA Hand Drawn Smoke Plumes}}
            \label{fig:noaa}
        \end{subfigure}
        \caption{\footnotesize{A selection of our available datasets, defined over the same region of British Columbia, Canada. Note that in \cref{fig:frp} we detail the size of our 10 km $\times$ 10 km grid cell at the asterisk. These individual layers are stacked together to compose a wildfire ``image'', with channels representing spatial distributions of relevant features for our \PM predictive task.}}
        \label{fig:datasets}
\end{figure}

BlueSky Canada is a similar air quality modeling framework based off a system pioneered by the U.S. Forest Service. BlueSky Canada offers hourly forecasts of \PM concentrations from forest fire up to $60$ hours in advance at a $4$ km resolution \citep{larkin2010bluesky}. It contrasts most significantly from FireWork in that contributions to \PM from pollution sources outside of wildfires are not incorporated \citep{larkin2010bluesky}.

\subsubsection{Meteorological Data}

The U.S. National Aeronautics and Space Administration (NASA) maintains Moderate Resolution Imaging Spectroradiometer (MODIS) instruments via their Terra and Aqua satellites, providing various quantifications of atmospheric variables at semi-regular intervals \citep{chudnovsky2014high}. Here, we are interested in the Aerosol Optical Depth (AOD) metric, which is available roughly every $6$ hours in a $24$ hour period. Wildfires generate a plethora of aerosol in dust, ash and smoke byproducts, which have a measurable effect on AOD. 
Previous work has also shown that AOD is a meaningful proxy for \PM \citep{chudnovsky2014high}.

We also include meteorological information from the NASA Modern Era Retrospective Analysis for Research and Applications, Version 2 (MEERA-2) program. Similar to \citep{yao2018machine}, we include eastward and northward components of wind vectors 50m above the surface, and at the 250 hPa and 500 hPa pressure levels, with a spatial resolution of $0.5\degree \times 0.625\degree$ (latitude $\times$ longitude). These variables have also held high predictive power in other wildfire smoke dispersion studies \citep{reid2015spatiotemporal}.

\subsubsection{Wildfire Data}

The MODIS instruments also provide data on fire locations and intensity. Intensities are approximated by the fire radiative power (FRP) variable, and fire locations are specified by a weighted centroid localization of FRP values in all $1$-km $\times$ $1$-km fire pixels (as determined by the active fire product). While \citep{yao2018machine} transform FRP to extract additional predictive variables, we rely on the expressivity of our highly non-linear network to learn potentially better features strictly from FRP.

Additionally, we include direct observations of smoke plumes from wildfires. The U.S. National Oceanic and Atmospheric Administration (NOAA) maintains a Hazard Mapping System (HMS) which dynamically assesses fire and smoke products. Analysts hand draw smoke plume boundaries based on available data from various fire detection sources \citep{stein2015noaa}. Due to the manual nature by which this dataset is acquired, significant temporal and spatial gaps occur. Nonetheless, this is the only component of our dataset which explicitly contains human confirmation of wildfire smoke behavior, and we therefore include it when available as a potential regularizer for other data sources.  We treat the smoke plumes as a binary variable over a predefined grid; 0 indicating smoke is not present, and 1 indicating smoke is present.

\subsubsection{Particulate Matter Ground Truth}

For ground truth \PM values, we use the $2018$ and $2019$ $1$-hour average \PM measurements from $56$ air quality monitoring stations through British Columbia (courtesy of the British Columbia Ministry of Environment and Climate Change Strategy).\footnote{Please refer to the supplementary material for a visualization of these stations within our grid of interest}. We log transform the values as per \citep{yao2018machine} to address its heavy right-skewed distribution.

\subsection{Multidimensional Wildfire Composition Images}


In order to amalgamate these varied data sources into a temporally and spatially consistent format which can be consumed by our convolutional neural network, there needs to be a significant amount of pre-processing done.

First, we define a regular, approximately square grid over the province of British Columbia, Canada, with (latitude, longitude) corners at (57.87, -133.54), (47.31, -127.18), (60.61, -112.19), (49.43, -110.61), and a grid resolution of $10$-km $\times$ $10$-km cells covering the roughly $1250$-km $\times$ $1250$-km area. This will serve as an image-like canvas on which we can populate different pixels with the requisite features.

Second, note that we seek to make predictions $24$-hours in advance. We extract temporal and spatial (latitude/longitude) labels for each element of each dataset. Then, for each time where we have \PM ground truths, we project available datapoints from all datasets $24$-hours prior to their corresponding cells and channels within our defined grid. Cells and channels for unavailable measurements are simply set to the closest available measurements from an earlier time, or to $-1$ when even those are absent. We also do a similar projection for the $56$ available \PM ground truths.

The final result is a $125 \times 125 \times 9$-dimensional input image for each timepoint of predictive interest, and a corresponding sparse $125 \times 125 \times 1$-dimensional output image of \PM labels corresponding to $24$-hours thereafter. In total, we train on $4870$, validate on $610$ and test on $610$ such input-output pairs, randomly shuffled.

\subsection{Model Architecture}

While a standard fully convolutional neural network such as DenseNet or UNet can consume our defined input and output the \PM map of interest, problems of extreme sparsity remain \citep{drozdzal2016importance, iandola2014densenet}. For any particular prediction, we have at most 56 ground truth \PM values in a grid of size $125 \times 125$. We therefore robustify our convolutional neural network by introducing sparsity invariance into its composite layers, and defining additional tasks which might offer dense learning signal.

\begin{figure}[ht]
\centering
    \includegraphics[width=.9\textwidth]{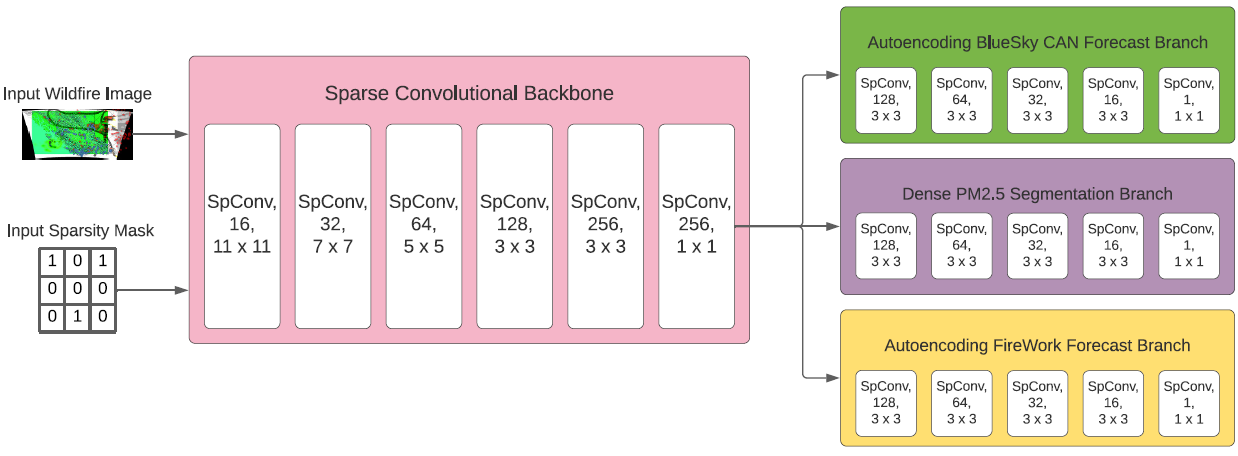}
    \caption{\footnotesize{Multiheaded model architecture. Each SpConv layer involves a sparse convolution layer as described in with the indicated kernel size and number of filters, with average pooling of the sparsity mask, followed by a non-linear activation (ReLU throughout, except the final layer in the individual task branches, where no activation is used.}}
    \label{fig:model}
\end{figure}

\subsubsection{Sparsity Invariant CNNs}

Sparsity invariant CNNs were developed by \cite{uhrig2017sparsity} as a means to preserve sparsity constraints throughout all layers of a convolutional neural network by explicitly accounting for a binary mask describing the sparsity pattern. In particular, such a ``sparse'' convolution involves pooling only over observed pixels of the image, and normalizing according to the mask. Here, our sparsity mask describes the locations of the available ground truth \PM values over the spatial grid. While \citep{uhrig2017sparsity} use max pooling to downsample the binary mask after each sparse operation, we find that average pooling leads to smoother inpainting of the resultant \PM output map. We employ these sparsity invariant layers in a core feature extraction backbone, as seen in \cref{fig:model}.

\subsubsection{Multitask Learning}

In an effort to maximally use smaller datasets, it is often helpful to fuse information from training signals of related tasks. The idea is that by sharing representations used for similar tasks, we can enable the model as a whole to generalize more efficiently to such tasks, and crucially, to our ultimate task of interest. This idea is known as multitask learning, or joint learning.

Here, the additional tasks we define are autoencoding ones; we define separate branches as in \cref{fig:model} which each output a $125 \times 125 \times 1$ map, consistent with the \PM forecasts of the FireWork and BlueSky Canada models (recall that these are incorporated as part of the input). While our limited corpus of highly localized wildfire data is one impetus for multitask learning, these autoencoding heads reveal another: we hope to provide learning signal where ground truths are unavailable, allowing the model to overcome extreme sparsity issues by borrowing from the learned dynamics contained within the baseline forecasting models. Then our final model loss $\mathcal{L}$ is defined as:

\[
\mathcal{L} = \gamma_1 ||I_{fw} - \hat{I}_{fw}||_1 + \gamma_2 ||I_{bscan} - \hat{I}_{bscan}||_1 + \gamma_3 ||I_{pm25} - \ M \odot \hat{I}_{pm25}||_1
\]

where $\gamma_{1, 2, 3}$ are hyperparameters, $I$ is the target map, $\hat{I}$ is the predicted map from the corresponding branch, with subscripts $fw$, $bscan$, $pm25$ denoting the FireWork baseline, BlueSky Canada baseline and \PM ground truths, respectively. $M$ denotes the binary mask demarcating the sparsity pattern of the \PM ground truths.

\section{Results and Discussion}

Given the geographical and temporal specificity of our approach, as well as the collective range of datasets used, there are no statistical models which offer direct comparisons. We therefore assess performance of our model by comparing model \PM predictions with FireWork and BlueSky Canada \PM predictions at the $56$ air monitoring stations over timepoints within the defined test set. We then also look at heatmaps of model predictions to ascertain whether or not meaningful interpolations are made in regions where ground truths are not available.

\begin{table}[ht]
\centering
\caption{We report average L1 error for our model and available baselines for early, mid and late temporal subsets of the test set. Lower is better, bolded is best.}
\label{tab:model-results}
\scalebox{.8}{
\begin{tabular}{lccc}
\toprule
	&\multicolumn{3}{c}{{Mean Absolute Error ($\mu g / m^3)$}} \\[1pt]
	\multicolumn{1}{c}{} & \makecell{\textbf{Early} \\ (Apr + May)} & \makecell{\textbf{Mid} \\ (Jun + Jul + Aug)} & \makecell{\textbf{Late} \\ (Sep + Oct)} \\

\midrule

\kw{FireWork} & 10.08 & 23.39 & 17.52 \\

\kw{BlueSky Canada} & 22.34 & 75.73 & 44.10  \\

\midrule
\kw{Our Model} & \textbf{2.26} & \textbf{14.21} & \textbf{6.79}  \\

\bottomrule
\end{tabular}
}
\end{table}

\cref{tab:model-results} details our model performance against the described baselines. Because \PM values can be dramatically higher during the peak of the fire season in July and August, we separate this assessment for different sets of months. In particular, we demarcate the early, mid and late fire seasons as April to May, June to August, and September to October, respectively. Note that we outperform both FireWork and BlueSky Canada at all points in the fire season, validating our residual learning approach and verifying that additional raw data is semantically useful for our model in improving the baseline predictions.

\begin{figure}[ht]
\centering
    \includegraphics[width=\textwidth]{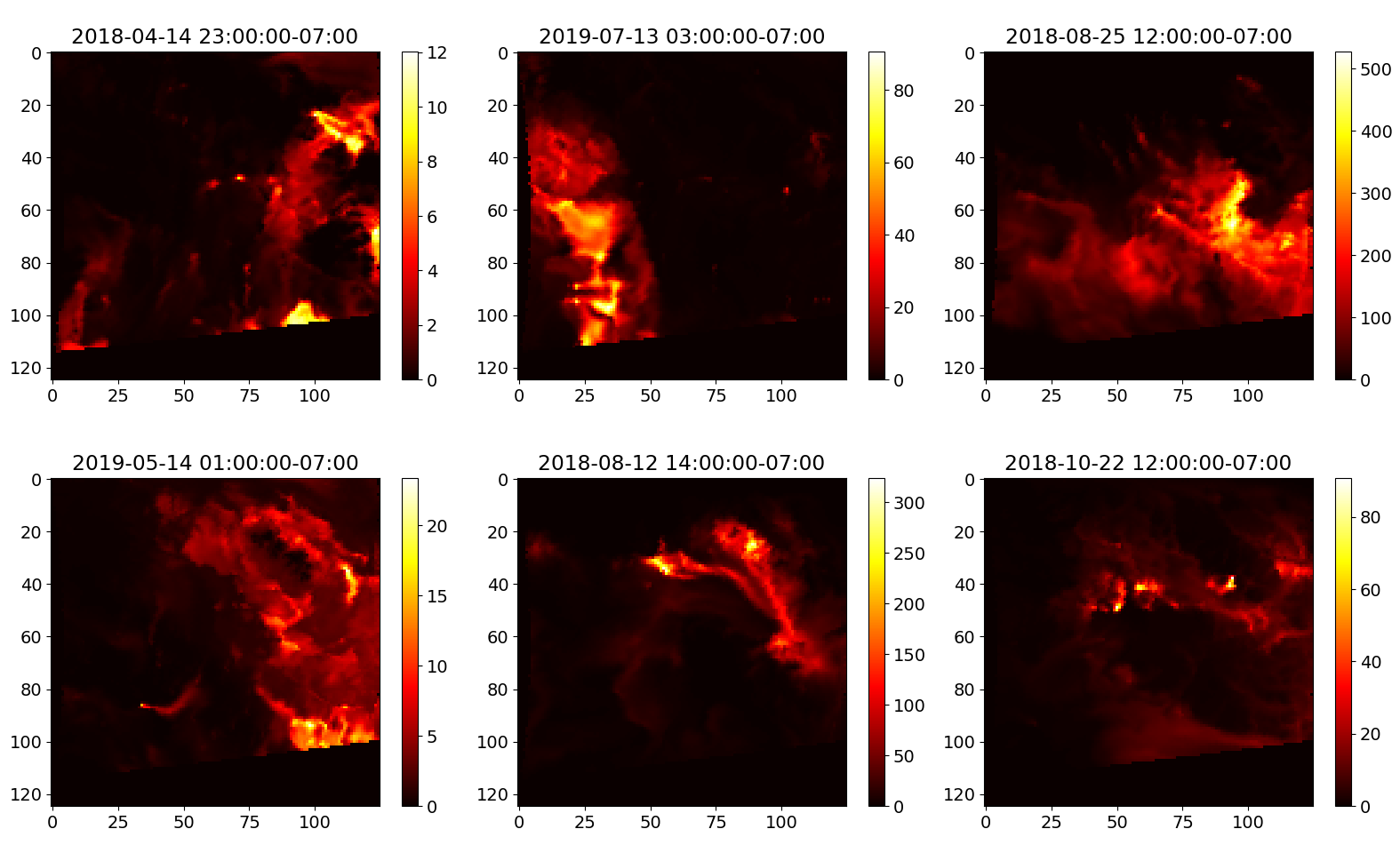}
    \caption{\footnotesize{Predictions made using available data 24 hours in advance. We note the smooth inpainting of \PM values despite lacking ground truths on nearly 99.7\% of the image. In future work we verify the implicit smoke dynamics within these predictions.}}
    \label{fig:heatmaps}
\end{figure}

In \cref{fig:heatmaps}, we show model predictions over the entire defined grid. Each pixel within each heatmap represents a $100$ $km^2$ area, and the heatmap resolution is $125 \times 125$ pixels. Firstly, note that we accurately capture highs and lows of \PM in correspondence with the fire season (beginning in April, peaking in July and August, and ending in October). Secondly, we see that despite lacking ground truth \PM values in between air monitoring stations, our model is able to interpolate \PM meaningfully, representing complex and diverse \PM falloff patterns and interactions between smoke dispersion from various fires. More work clearly needs to be done to verify these implied dynamics, and we leave this for the future. \footnote{Please refer to the supplementary material for additional prediction frames.}

\section{Conclusion}

In this work, we tackle the challenging but important task of forecasting the public health burden of smoke particulate matter perpetuated by wildfires. By incorporating baseline forecasting models and raw meteorological and wildfire data variables from satellite measuring systems, we are able to design an input data format which preserves spatial relationships. We then overcome sparsity issues plaguing more traditional statistical modeling frameworks by introducing sparsity invariant layers, and defining auxiliary tasks that provide guiding intermediate learning signal to the network. We demonstrate strong results on real world wildfires as compared to forecasting systems currently in use. Future work will ascertain that our method is generalizable outside British Columbia, Canada, and that the smoke particulate matter behaviors modeled by our network are consistent with domain expectations.

\begin{ack}

We would like to thank the British Columbia Centre for Disease Control (BCCDC) for their invaluable support in acquiring and cleaning datasets, and their direction to previous relevant work. In particular, Dr. Sarah Henderson and her team, Kathleen McLean, Naman Paul and Angela Yao were instrumental in identifying predictive variables for our requisite tasks. We would also like to acknowledge the University of British Columbia Data Science Institute (DSI) and its partners; RW, AB, BDR and MT were all supported by DSI's Data Science for Social Good (DSSG) program during the tenure of this work, with compute credits provided by Microsoft. Finally, we would like to thank the support of the Canadian Institute for Advanced Research (CIFAR).

\end{ack}

\bibliographystyle{plainnat}
\bibliography{refs}

\end{document}